\documentclass[letterpaper, 10pt, conference]{ieeeconf}  

\usepackage{times}
\usepackage{epsfig}
\usepackage{graphicx}
\usepackage{amsmath}
\usepackage{amssymb}
\usepackage{xcolor}
\usepackage{subcaption}
\usepackage[breaklinks=true,bookmarks=false]{hyperref}

\usepackage{balance}
\DeclareMathAlphabet{\pazocal}{OMS}{zplm}{m}{n}
\newcommand{\Lb}{\pazocal{L}}
\IEEEoverridecommandlockouts

\title{\LARGE \bf Bayesian Deep Neural Networks for Supervised Learning of Single-View Depth}

\author{Javier Rodríguez-Puigvert, Rubén Martínez-Cantín, Javier Civera
\thanks{The authors are with I3A, Universidad de Zaragoza, Spain.
        {\tt\small \{jrp,rmcantin,jcivera\}@unizar.es}}
}

\begin{document}

\maketitle
\thispagestyle{empty}
\pagestyle{empty}

\begin{abstract}
Uncertainty quantification is essential for robotic perception, as overconfident or point estimators can lead to collisions and damages to the environment and the robot. In this paper, we evaluate scalable approaches to uncertainty quantification in single-view supervised depth learning, specifically MC dropout and deep ensembles. For MC dropout, in particular, we explore the effect of the dropout at different levels in the architecture. 

We show that adding dropout in \emph{all} layers of the encoder brings better results than other variations found in the literature. This configuration performs similarly to deep ensembles with a much lower memory footprint, which is relevant for applications. Finally, we explore the use of depth uncertainty for pseudo-RGBD ICP and demonstrate its potential to estimate accurate two-view relative motion with the real scale.

\end{abstract}

\section{Introduction}

The quantification of the uncertainty is critical in robotics, in order to implement systems that are robust and reliable in real-world applications. Point estimators, which dominate the landscape of multi-view~\cite{engel2017direct,mur2017orb} and single-view~\cite{Fu2018,Godard2018} scene reconstruction, do not typically compute the full distribution but just the maximum-likelihood state. Higher-level decision blocks have no means to judge how accurate these estimates are, and hence its use for safe planning might be questionable. Uncertainty is often present in the formulation of model-based estimators (e.g., \cite{barfoot2017state}), but much less in the training of deep learning models. Furthermore, learning-based approaches tend to overfit on standard datasets, which might lead us to assume a reasonable general performance while they are strongly biased. In such cases, their outputs should not be trusted in real-world applications, for which we would need generalizable and self-aware models. Bayesian learning is one of the approaches that can address such application challenges.

\begin{figure}[ht!] 
    \centering
    \includegraphics[width=0.85\columnwidth]{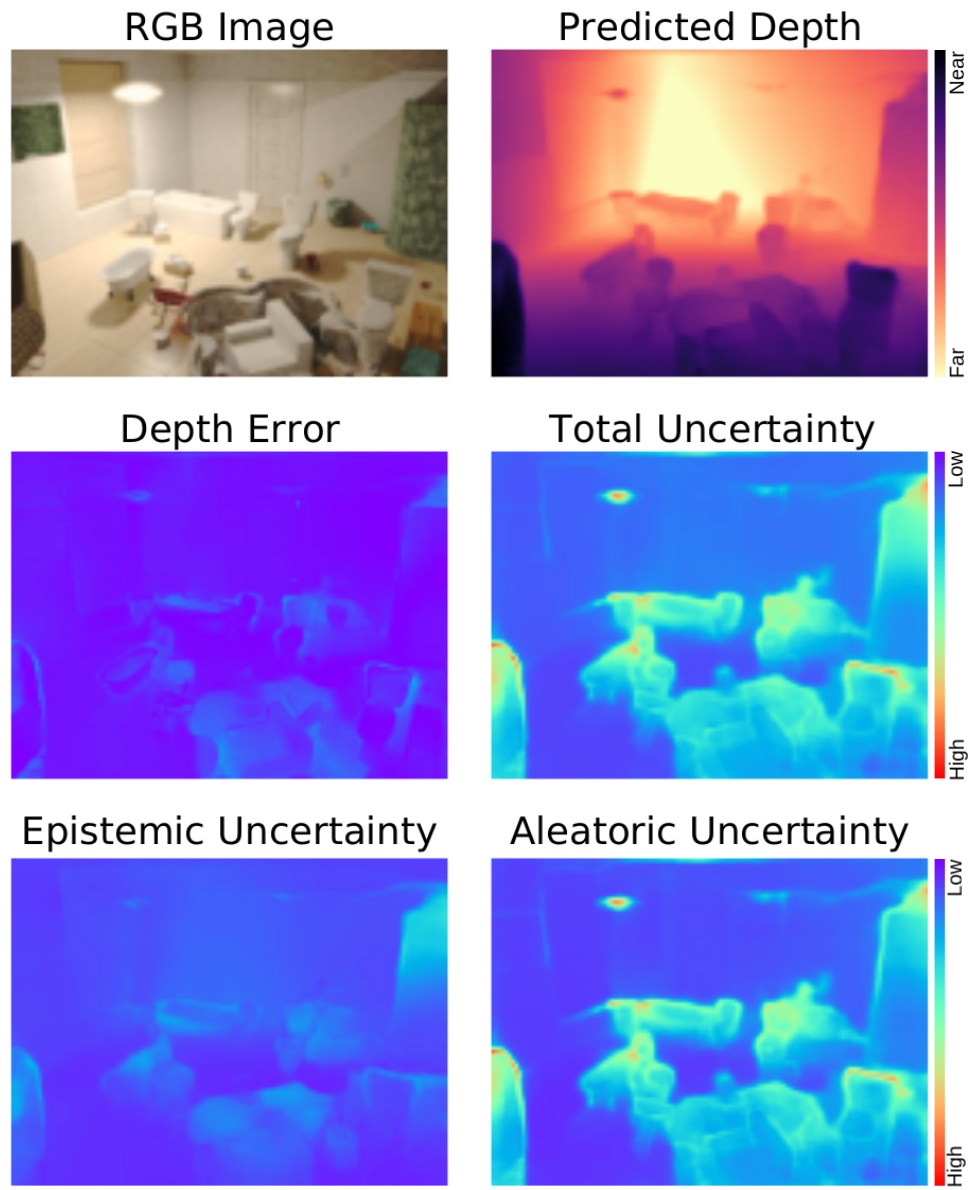}
    \caption{Bayesian single-view depth prediction for a SceneNet image. Notice in the middle row the small depth error, and how the total uncertainty models it accurately. Notice also in the bottom row how epistemic and aleatoric sources are both significant and relevant for uncertainty quantification.}
    \label{fig:teaser}
\end{figure}

In this work, we evaluate two sources of uncertainty --epistemic and aleatoric-- in supervised depth learning, with the aim to determine the quality of the model predictions and hence its potential for robotic applications. In neural networks, uncertainty can stem from the input data or the network weights. For the latter, scalable approaches to Bayesian deep learning have shown to be effective to model uncertainty. Fig.~\ref{fig:teaser} shows the output of our Bayesian framework for supervised single-view depth learning. Notice first the small depth error, but most importantly, how such error is mostly coherent with the predicted uncertainty. Observe also how the aleatoric uncertainty is in this case bigger, but the epistemic one is still relevant for an accurate quantification.

The contributions of this paper are: first, we provide a unified framework and a thorough evaluation of scalable uncertainty estimation methods, namely Monte Carlo (MC) dropout and deep ensembles, for supervised single-view depth learning with deep convolutional networks. 
Secondly, we propose to apply MC dropout in the encoder, contrary to recent works~\cite{Gustafsson2019,poggi2020uncertainty} that apply it in the decoder. We demonstrate in our evaluations that, in the particular task of supervised depth learning, the dropout in the encoder achieves a better performance than the dropout in the decoder.

Such a result has relevant practical implications, as MC dropout has a lower memory footprint than ensembles. Finally, we also provide results in pseudo-RGBD ICP, a potential application for our single-view depth uncertainty models. In our experiments, we demonstrate that our uncertainty estimates are reasonably well calibrated and has significant potential to provide accurate and scaled motion estimates from monocular views.

\section{Background and Related Work}
\subsection{Structure Estimation and Learning from Images}

Reconstructing a 3D scene from visual data has been addressed from a wide variety of perspectives, the more typical being based on multiple images either alone~\cite{mur2017orb,engel2017direct} or fused with other sensors (e.g., visual-inertial setups~\cite{qin2018vins}). However, multi-view and visual-inertial pipelines present two limitations: they require sufficiently textured scenes to find correspondences, and also sufficient motion for observability. Single-view depth learning can help with these two issues, although it is significantly more challenging due to its ill-posed nature.

Following the seminal work of \cite{make3d} on single-view depth, Eigen et al. \cite{Eigen2014} were the first ones that used depth supervision to train a deep network for such task. 
Many works followed with different contributions: Laina et al. \cite{Laina2016} proposed deeper fully convolutional models. Fu et al. \cite{Fu2018} proposed a spacing-increasing depth discretization that learns depth from an ordinal regression perspective. Dijk and de Croon \cite{dijk2019neural} evaluated a self-trained networks to investigate which visual cues they use. From their conclusions, depth networks favor vertical positions and disregard obstacles in their apparent size. Similarly, our work contributes to understanding the behavior of depth networks from a Bayesian perspective.  

Several works have proposed self-supervised approaches, using photometric reprojection losses between stereo or multiple views~\cite{Godard2017,Zhou2017}. However, self-supervised approaches still underperform compared to supervised ones. For this reason, and given the current abundance of RGB-D data and the huge growth and potential of such sensors, we focus on supervised methods in this work.

\subsection{Bayesian Deep Learning}
\label{sec:bayesdl}
Bayesian deep learning combines the strengths of deep neural architectures with the uncertainty quantification of probabilistic (Bayesian) learning and inference methods. Regarding uncertainty, we must differentiate between what the model does not know and what is missing from the input data. Accordingly, the uncertainty sources can be classified into two: aleatoric and epistemic. 
Aleatoric (also referred to as statistical) uncertainty, refers to the variations caused by the realization of different experiments with stochastic components. In our models, it encodes the variability in the different inputs from the test data and hence cannot be reduced by increasing the amount of training data. Some models assume that the aleatoric uncertainty is homoscedastic; that is, it is independent of the input data. In this work, we train the network to predict the uncertainty for each input datum resulting in a heteroscedastic uncertainty model~\cite{Kendall2017, Kiureghia}.

Epistemic (also known as systematic) uncertainty, represents the lack of knowledge of a trained model. This type of uncertainty is deeply related to the training data and the model ability to generalize. For example, epistemic uncertainty is high for out-of-distribution data or extrapolation in regions where training data was scarce. In Bayesian deep learning, the epistemic uncertainty can be estimated from the uncertainty in the model parameters, assuming that the model architecture is correct. In this case, epistemic uncertainty can be obtained using a prior distribution $p(\mathbf{\theta})$ over the neural network parameters $\mathbf{\theta}$ and computing its posterior distribution $p(\theta|\mathcal{D})$ given a dataset $\mathcal{D}$ using Bayes rule: $p(\theta|\mathcal{D}) =\frac{p(\mathcal{D}|\theta)p(\theta)}{p(\mathcal{D})}$. In general, this equation is intractable for state-of-the-art deep architectures, but there are several approaches to tackle this problem that we describe below. 

\textbf{Variational inference (VI).} VI proposes the use of a tractable approximation $q(\theta)$ to the posterior distribution $p(\theta|\mathcal{D})$. Mean-field variational inference assumes an isotropic Gaussian distribution for $q(\theta)\sim \mathcal{N}(\theta|\mu,\mathbf{I}\sigma)$. The parameters of the approximate distribution $q(\theta)$ are optimized by minimizing the KL-divergence between the approximate distribution and the true posterior $D(q||p)$. Mean-field variational inference with Gaussian approximation suffers from the soap-bubble effect, reducing the predictive performance as most samples fall in a ring. The Radial Bayesian Neural Networks \cite{farquhar_radial_2020} avoid that effect, but the distribution is biased towards the center resulting in uncertainty underestimation. Furthermore, VI methods are very sensitive to calibration and configuration. A natural-gradient VI method \cite{Osawa2019} was introduced to improve the robustness of the optimization. However, it requires strong approximations of the Hessian, resulting in lower performance.

\textbf{Monte Carlo (MC) Dropout.} MC dropout can be used to approximate the posterior distribution, as proposed in \cite{Gal2016}. It can be considered as a specific case of VI, where the variational distribution includes a set of binary random variables that represent the corresponding unit to be turned off or dropped. The approximation makes the computation more tractable and robust. MC dropout is able to approximate multimodal distributions. However, the epistemic distribution on the weight-space only has discrete support. \cite{Kendall2017} presented a framework to combine both aleatoric and epistemic uncertainty, where MC dropout is used to obtain epistemic uncertainty, while the function mapping the aleatoric uncertainty is learned from the input data. 


\textbf{Deep Ensembles.} Deep ensembles \cite{Lakshminarayanan2017} involves training the same architecture many times optimizing some MAP loss, but starting from different random initialization of its parameters. Therefore, deep ensembles are not truly a Bayesian approach as the samples are distributed according to the different local optima. Conversely, these models in an ensemble perform reasonably well, even considering the small number of random samples considered in practice, as all of the models are optimized and have a high likelihood. Therefore, deep ensembles can be considered an approximate Bayesian model average, although, in practice, they can also be used as a rough posterior approximation. Contrary to MC dropout, where the model weights are shared between samples, in the case of deep ensembles, each \emph{sample} is trained independently. Therefore the number of model parameters required grows linearly with the number of samples. Furthermore, deep ensembles also result in a distribution with discrete support on the weight-space.  

\subsection{Bayesian Deep Learning in Computer Vision}

Evaluating uncertainty correctly is still in open discussion, as it is task-related. Mukhoti and Yal \cite{Mukhoti2018} evaluated MC dropout for semantic segmentation and designed the metrics for such case. Similarly, Gustafsson et al. \cite{Gustafsson2019} designed a framework to explore uncertainty metrics for semantic segmentation and depth completion, using MC dropout and deep ensembles. 


Ilg et al. \cite{Ilg2018} compare different strategies and techniques for quantifying the uncertainty of optical flow. They also introduce a multi-hypothesis network based on a winner-takes-all loss function that penalizes the best hypothesis result. However, disentangling aleatoric and epistemic uncertainty could be an arduous task in multi-hypothesis approaches. The network that merge all hypothesis contains its own epistemic uncertainty that is not taking into account. Nevertheless, they show to be competitive in comparison to deep ensemble and MC dropout. For depth estimation, Yang et al. \cite{yang2019inferring} proposes a multinomial distribution to learn uncertainty based on discretizing the depth space.

For our case of single-view depth regression, Poggi et al. \cite{poggi2020uncertainty} evaluated the uncertainty in self-supervised networks, which leverage photometric consistency between views. They observed that depth accuracy is improved by uncertainty estimation along the training paradigms. Our work complements the ones mentioned in this section by evaluating uncertainty quantification in a supervised regression setting. 

\section{Bayesian Single-View Depth Learning from Supervised Data}

As described in Section \ref{sec:bayesdl}, MC dropout and deep ensembles provide a sample representation of the posterior distribution over the network parameters. Here, we introduce a unified formulation to analyze the posterior and predictive distribution for these sample representations. For brevity, we have particularized our framework for depth perception applications, although it could be extended to other tasks.

\subsection{Architecture and Loss}

We adapt a U-Net \cite{Ronneberger} encoder-decoder architecture as in  \cite{Godard2018,poggi2020uncertainty}. Our encoder is a Resnet18 \cite{He2015} pre-trained in ImageNet~\cite{Russakovsky2014}. Table \ref{tab:decoder} summarizes our decoder architecture. 

\begin{table}[h!]
\centering
\scriptsize
\begin{tabular}{ |l|l|l|l|l| }
\hline
\multicolumn{4}{ |c| }{Depth Decoder} \\
\hline
\textbf{layer} & \textbf{\# filters} & \textbf{inputs} & \textbf{activation} \\ \hline
upconv5 & 256 & econv5 & ELU\\
iconv5 & 256 & ↑upconv5, econv4 & ELU \\\hline
upconv4 & 128 & iconv5 & ELU \\ 
 iconv4 & 128 & ↑upconv4, econv3 & ELU \\
depth\_unc4 & 2 &  iconv4 & - \\ \hline
upconv3 & 64 & iconv4 & ELU\\
iconv3 & 64 & ↑upconv4, econv3 & ELU \\ 
depth\_unc3 & 2  &  iconv3 & - \\\hline
upconv2 & 32 & iconv3 &ELU\\
iconv2 & 32& ↑upconv2, econv1  &  ELU\\
depth\_unc2 & 2 & iconv2 & - \\\hline
upconv1 & 16 & iconv2 & ELU\\
iconv1 & 16 & ↑upconv1 & ELU\\
depth\_unc1 & 2 & iconv1& -\\
\hline
\end{tabular}
    \caption{Decoder architecture. Kernels are always $3\times 3$ with stride $1$. ↑ stands for $2 \times 2$ nearest-neighbor upsampling.}
    \label{tab:decoder}
\end{table}

Our training data $\mathcal{D}=\{ \{\mathcal{I}_1,d_1 \},\hdots,\{\mathcal{I}_N,d_N \}\}$ is composed by $N$ supervised pairs, each pair $i \in \{1,\hdots,N\}$ containing a RGB image $\mathcal{I}_i \in \{0,\hdots,255\}^{w \times h \times 3}$ and its ground truth depth $d_i \in \mathbb{R}^{w \times h}_{>0}$. For a single input image, the network $f_\theta(\mathcal{I})$ outputs two channels: per-pixel depth $\widehat{d}(\mathcal{I})$ and uncertainty $\sigma_d(\mathcal{I})$. The later corresponds to aleatoric uncertainty, which can also be interpreted as heteroscedastic observation noise. 
We incorporate both output channels in a single loss per image by using a standard Laplace log-likelihood \cite{Kendall2017}:

\begin{equation}
    \Lb(\theta) = \frac{1}{w \cdot h} \sum_{j \in \Omega} \frac{||d_j - \widehat{d}(\mathcal{I}_j)||}{ \sigma_{d}(\mathcal{I}_j)} + \log \sigma_{d}(\mathcal{I}_j)
\end{equation}
where $j \in \Omega$ is the pixel index in the image domain $\Omega$.

For deep ensembles, the loss function is evaluated independently for each sample model as they are trained separately, resulting in $M$ sets of parameters $\{\theta_m\}_{m=1}^M$. Although the sample models are not drawn from the posterior distribution, it still can be considered an approximation in practice. Deep ensembles are especially suitable for our problem as we need to maintain the number of samples small to keep it tractable. Therefore, it is important that we are not wasting valuable resources in low probability models that might reduce the overall performance.

In the case of MC dropout, the loss function can be used for approximate variational inference on the posterior distribution of the weights by training with dropout {after } every layer. The actual Monte Carlo phase is done by also performing random dropout at test time to sample from the variational distribution computed during training \cite{Kendall2017}. This sampling at test time results again in a set of $\{\theta_m\}_{m=1}^M$ different parameters. However, note that this time they are all generated from the same trained model, resulting in a much lower memory and computational footprint compared to deep ensembles or other variational methods. 

In practice, we found that adding dropout at every layer reduced the predictive performance considerably for our application, which is consistent with previous results \cite{Mukhoti2018}. Therefore, in section \ref{sec:results-bayesdepth}, we study different configurations of dropout and compare their quality both in terms of depth error and uncertainty quantification.

\subsection{Bayesian prediction of sample-based deep networks}
The predictive distribution for a pixel depth can be computed by integrating over the model parameters. We use the same strategy for MC dropout and deep ensembles, as they both use sample representations of the model parameters: 
\begin{equation}
    \begin{split}
        p(d | \mathcal{I}, \mathcal{D} ) &= \int p(d | \mathcal{I}, \theta) p(\theta|\mathcal{D}) d\theta\\
        & \approx \sum_{m=0}^M p(d | \mathcal{I}, \theta_m)
    \end{split}
\end{equation}

As our architecture generates a Gaussian prediction for the pixel depth $\mathcal{N}(\widehat{d},\sigma^2_d)$, the sample-based output is a mixture of Gaussians that can be approximated by a single Gaussian. In particular, for the $\{\theta_m\}_{m=1}^M$ model samples (MC Dropout) or models (ensembles) with respective outputs $\widehat{d}(m)$ and $\sigma_d(m)$, we approximate the total predictive distribution per pixel as a Gaussian $p(d | \mathcal{I}, \mathcal{D}) \approx \mathcal{N}(\widehat{d}_t,\sigma^2_t)$ with:
\begin{equation}
    \begin{split}
        \widehat{d}_t &= \frac{1}{M} \sum_{m=1}^{M} \widehat{d}_m\\
        \sigma^2_t &= \underbrace{\frac{1}{M} \sum_{m=1}^{M} \left(\widehat{d}_t - \widehat{d}(m)\right)^2}_{\text{epistemic}} + \underbrace{\frac{1}{M} \sum_{m=1}^{M} \sigma^2_d(m)}_{\text{aleatoric}}
    \end{split}
\end{equation}

In the experiments, we will show that identifying and quantifying the epistemic from the aleatoric uncertainty will be fundamental to finding the uncertainty source and improving the quality of the model and the predictions.

\section{Experiments}

\begin{figure}
    \centering
    \centering
    \includegraphics[width=\columnwidth]{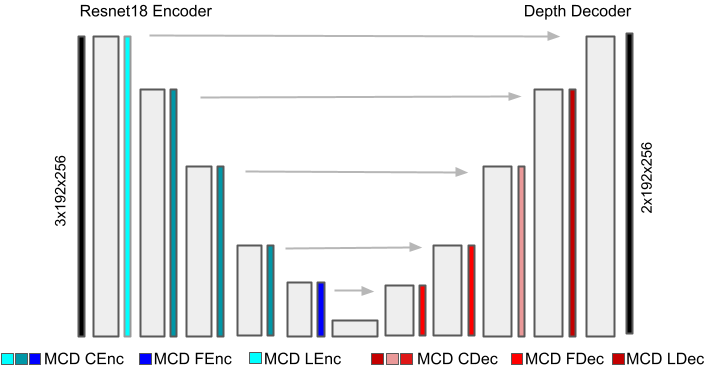}
    \caption{Variations of MC Dropout in our experiments.  
    }
    \label{fig:arch}
\end{figure}

\begin{table*}
\scriptsize
    \centering
    \begin{tabular}{|c|c||c|c|c|c|c|c|c||c|c|}
    \hline
        Model & M & Abs Rel  & Sq Rel & RMSE & RMSE Log & $\delta< 1.25$ & $\delta < 1.25^2$ & $\delta < 1.25^3$ & AUCE  & AUSE\\\hline
MCD CEnc p=0.3 & 64 & \textbf{0.1231} & \textbf{0.1222} & \textbf{0.6396} & \textbf{0.1982} & \textbf{0.8719} & \textbf{0.9635} & \textbf{0.9850} & 0.4940 & 0.0985\\\hline
MCD CEnc p=0.5 & 64 & \underline{0.1243} & \underline{0.1228} & \underline{0.6484} & 0.2023 & \underline{0.8675} & \underline{0.9615} & 0.9842 & \underline{0.4833} & \underline{0.0968}\\\hline
        MCD FEnc p=0.3  & 64 & 0.1291 & 0.1326 & 0.6688 & 0.2047 & 0.8591 & 0.9591 & 0.9834 & 0.4954 & 0.1210\\\hline
MCD FEnc p=0.5 & 64 & 0.1320 & 0.1363 & 0.6633 & 0.2062 & 0.8583 & 0.9586 & 0.9833 & 0.4869 & 0.1229\\\hline
MCD LEnc p=0.5 & 64 & 0.1244 & 0.1245 & 0.6582 & 0.2010 & 0.8658 & 0.9607 & 0.9846 & 0.4955 & 0.1219\\\hline
MCD LEnc p=0.3 & 64 & 0.1244 & 0.1248 & 0.6769 & 0.2000 & 0.8629 & 0.9595 & 0.9848 & 0.4955 & 0.1327\\\hline
MCD CDec p=0.3 & 64 & 0.1316 & 0.1322 & 0.6781 & 0.2044 & 0.8567 & 0.9597 & 0.9841 & 0.4917 & 0.1268\\\hline
MCD CDec p=0.5 & 64 & 0.1369 & 0.1378 & 0.6988 & 0.2080 & 0.8494 & 0.9579 & 0.9835 & 0.4899 & 0.1323\\\hline
MCD FDec p=0.5 & 64 & 0.1263 & 0.1252 & 0.6690 & 0.2035 & 0.8604 & 0.9593 & 0.9841 & 0.4872 & 0.1353\\\hline
MCD FDec p=0.3 & 64 & 0.1264 & 0.1263 & 0.6622 & 0.2016 & 0.8641 & 0.9604 & 0.9842 & 0.4956 & 0.1304\\\hline
MCD LDec p=0.5 & 64 & 0.1336 & 0.1371 & 0.6866 & 0.2065 & 0.8568 & 0.9588 & 0.9833 & 0.4933 & 0.1241\\\hline
MCD LDec p=0.3 & 64 & 0.1293 & 0.1303 & 0.6782 & 0.2042 & 0.8589 & 0.9593 & 0.9837 & 0.4942 & 0.1285\\\hline
Deep ensembles & 18 & 0.1283 & 0.1244 & 0.6529 & \underline{0.1993} & 0.8617 & 0.9613 & \textbf{0.9850} & \textbf{0.4823 }& \textbf{0.0838}\\\hline

    \end{tabular}
    \caption{Depth and uncertainty metrics for several variations of MC Dropout and Deep Ensembles in SceneNet RGB-Depth. Best results are boldfaced, second best ones are underlined.}
    \label{tab:metrics}
\end{table*}
\begin{figure*}[ht!]
    \centering
    \includegraphics[width=\textwidth]{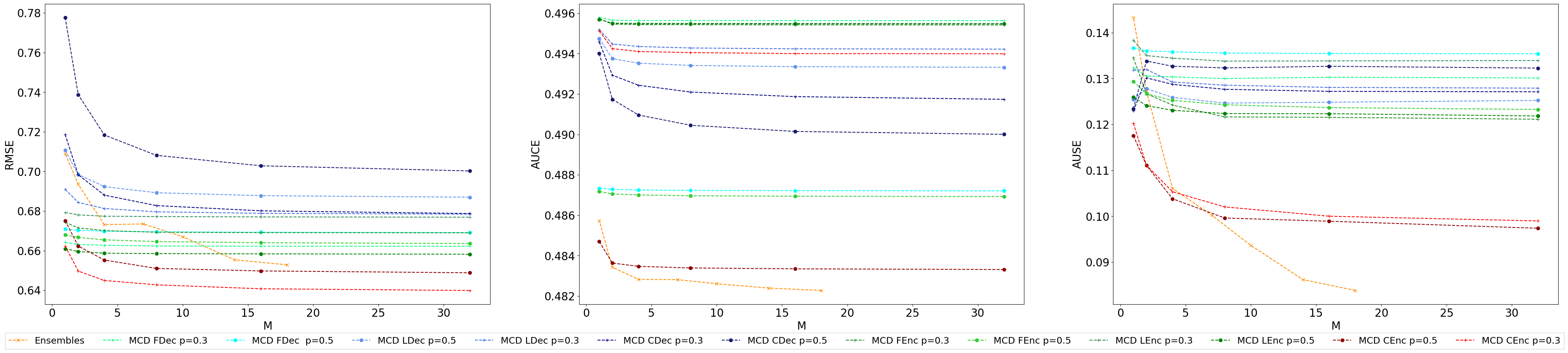}
    \caption{Comparison of MC dropout variations and deep ensembles for different numbers of forward passes $M$. Left: RMSE. Center: AUSE. Right: AUCE. The higher $M$ is, the better the performance, but with slight improvements for $M>18$.} 
    \label{fig:results}
\end{figure*}

\subsection{Datasets} 

\textbf{SceneNet RGB-D dataset~\cite{McCormac2016}} contains photorealistic sequences of synthetic indoor scenes from general camera trajectories, along with their ground truth. Our models are trained over $210,000$ synthetic images of $700$ scenes and tested on $90,000$ images of $300$ different scenes. We chose this dataset as it provides a wide variety of viewpoints and scenes, challenging occlusions and different lighting conditions, which are relevant for the network generalization.

\textbf{NYU Depth V2 \cite{Silberman:ECCV12}} consists of $120,000$ RGB-D images in $464$ indoor scenes. For training, we use $36,253$ images of $249$ scenes as proposed by  Lapdepth \cite{Lapdepth}. We test our models in the official split of $654$ images \cite{Eigen2014}. 

\subsection{Metrics}

We use the depth error metrics that are standard in literature: Absolute Relative difference, Square Relative difference, RMSE, RMSE log and $\delta < 1.25^i$ with $i \in \{1,2,3\}$ (see their definitions in \cite{Eigen2014}). For the pseudo-RGBD Bayesian ICP in Section \ref{sec:pseudorgbdicp}, we report the translational and rotational RMSE. For uncertainty, we use the Area Under the Calibration Error curve (AUCE) and Area Under the Sparsification Error curve (AUSE). Following Gustafsson et al. \cite{Gustafsson2019}, we define $100$ prediction intervals of confidence level $p \in \left[0,1 \right]$ and use the cumulative density function of our output distribution (Gaussian). For each confidence level, we expect for a perfect calibration that the ratio of the prediction interval covering the true target $\widehat{p}$ to be identical to the confidence level $p$. AUCE is defined as the area between the absolute error curve with respect to a perfect calibration $|p-\hat{p}|$.
As a second metric for the uncertainty quality, we use AUSE in terms of RMSE. Introduced by Ilg et al. \cite{Ilg2018}, it is a relative measure of uncertainty. This metric compares the ordering of the per-pixel uncertainties against the order of the per-pixel depth errors. The ordering should be similar for a well-calibrated uncertainty, as uncertain predictions will tend to have larger errors.

\subsection{Bayesian Single-View Depth}
\label{sec:results-bayesdepth}

We evaluate several variations of MC dropout and deep ensembles. Specifically, for MC dropout, we report depth and uncertainty metrics for dropouts at different layers and with $p = 0.3$ and $p = 0.5$, $p$ being the probability of an element to be zeroed.

\textbf{MC Dropout.} We consider seven variations (see Fig. \ref{fig:arch} for a summary plot):
In the decoder, dropout after every convolutional layer, (\textbf{MC D}ropout \textbf{C}omplete \textbf{Dec}oder), the first two convolutional layers (\textbf{MC D}ropout \textbf{F}irst \textbf{Dec}oder), and the last convolution layer of the decoder (\textbf{MC D}ropout \textbf{L}ast \textbf{Dec}oder). And, in the encoder, dropout after every convolutional layer (\textbf{MC D}ropout \textbf{C}omplete \textbf{Enc}oder), the first convolutional layer (\textbf{MC D}ropout \textbf{F}irst \textbf{Enc}oder), and after the last convolution layer (\textbf{MC D}ropout \textbf{L}ast \textbf{Enc}oder). Finally, in NYU RGB-D v2 we also evaluate the effect of the dropout in all layers of the architecture (\textbf{MC D}ropout \textbf{C}omplete \textbf{Enc}oder-\textbf{C}omplete \textbf{Dec}oder).

\textbf{Deep Ensembles.} We examine uncertainty and depth by averaging an ensemble composed of a variable number of networks. We initialise the network weights with different seeds from a normal distribution $\mathcal{N}(0,10^{-2})$.

\textbf{Results.} Table \ref{tab:metrics} shows the depth and uncertainty metrics for the MC dropout variations and deep ensembles  on SceneNet RGB-D.
We observe that the uncertainty metrics (AUCE and AUSE) for deep ensembles outperforms all variants of MC dropout. 
This is due to the fact that deep ensembles are optimized to be close to a minimum. 
However, the depth error metrics are consistently better for MCD CEnc, which also show the second best uncertainty metrics.

The performance of the different MC dropout models varies significantly, which is a novel result of our analysis. We found that introducing dropout in all layers of the encoder (MCD CEnc) improves the results with respect to applying it to a few layers of the encoder (MCD FEnc, LEnc) or to the decoder (MCD FDec, LDec and CDec). Again, this result is of relevance as MC dropout is commonly applied in the decoder for depth estimation \cite{Gustafsson2019,poggi2020uncertainty}. Our rationale for this result is as follows: we believe that applying MC dropout only in the decoder makes the network learn deterministic representations (the encoder is deterministic). In contrast, applying MC dropout in the image encoder allows us to learn probabilistic representations modeling uncertainty in the feature space. This seems to be a more appropriate choice for Bayesian image processing.

Our experiments show that there is small variations in the MC samples when we solely apply dropout close to the code, as done in MCD FDec or MCD Lenc. This results in worse calibrated models and also less satisfactory depth estimations. We also observed that applying MC dropout after the first and before the last layers of the network leads to poor performance (see MCD LDec, MCD FEnc).

Comparing the two dropout probabilities, $p = 0.3$ and $p = 0.5$, we observe that the depth prediction is usually better for $p = 0.3$ but the uncertainty is better calibrated for $p = 0.5$, as it introduces greater variability. One or the other should be preferred depending on the application.

MC dropout has a much smaller memory footprint than deep ensembles and our results indicate that the MCD CEnc performs similarly to ensembles, so it could be relevant option in certain applications. 
Specifically, only $56$ Mb are required for our MC dropout models, while the memory for deep ensembles grows linearly with the number of samples (for $M=18$, around $1$ Gb). The run time grows linearly with the number of samples in both cases (around 3--4 ms per forward pass). All data stems from a NVIDIA GeForce RTX 3090 GPU.

\begin{figure}
\centering
\begin{subfigure}{.45\columnwidth}
  \includegraphics[width=\linewidth]{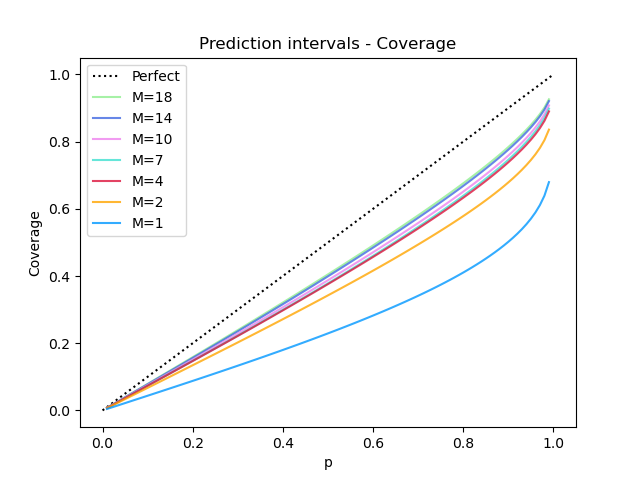}
  \caption{\scriptsize AUCE Deep Ensembles}
\end{subfigure}
\begin{subfigure}{.45\columnwidth}
  \includegraphics[width=\linewidth]{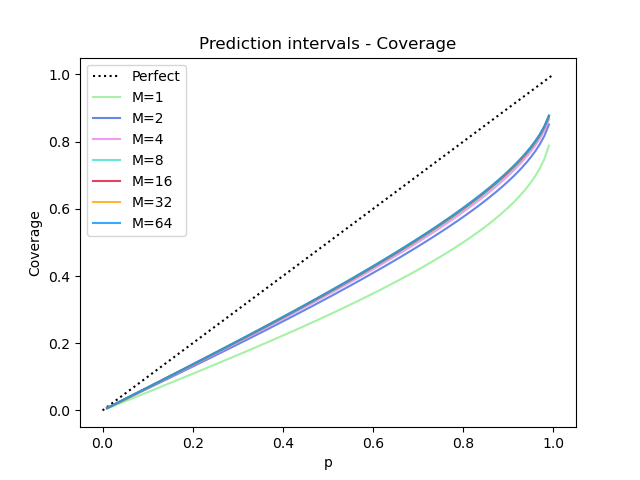}
  \caption{\scriptsize AUCE MCD CEnc p=0.5}
\end{subfigure}
\begin{subfigure}{.45\columnwidth}
  \includegraphics[width=\linewidth]{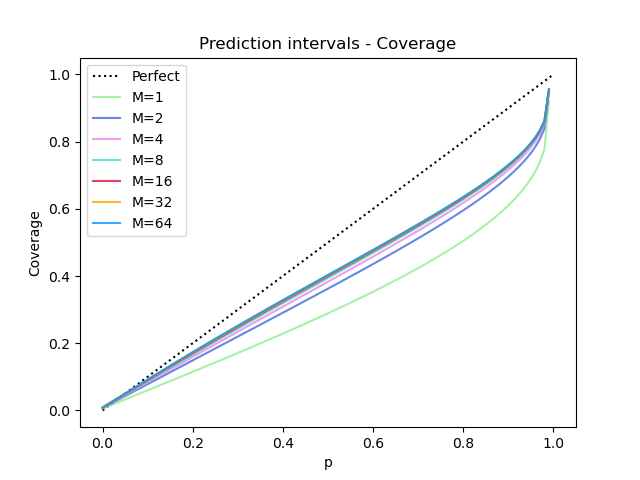}
  \caption{\scriptsize AUCE MCD CDec p = 0.3}
\end{subfigure}
\begin{subfigure}{.45\columnwidth}
  \includegraphics[width=\linewidth]{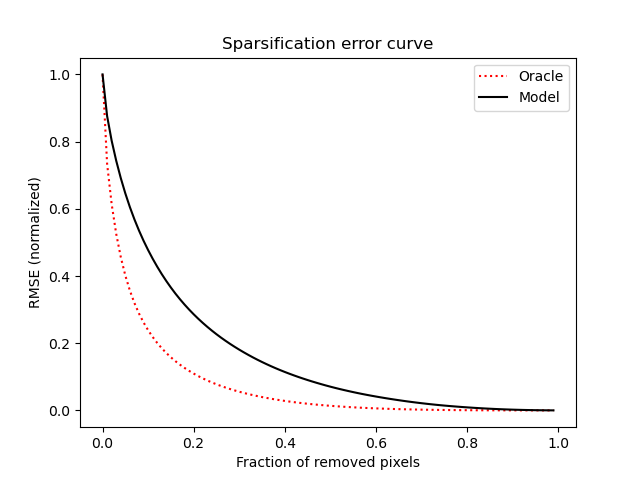}
  \caption{\scriptsize AUSE Deep Ensembles M=18}
\end{subfigure}
\begin{subfigure}{.45\columnwidth}
  \includegraphics[width=\linewidth]{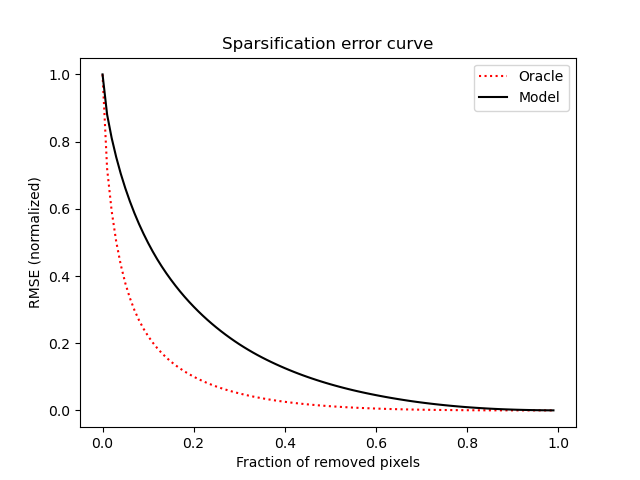}
  \caption{\scriptsize AUSE MCD CEnc p=0.5 M=64}
\end{subfigure}
\begin{subfigure}{.45\columnwidth}
  \includegraphics[width=\linewidth]{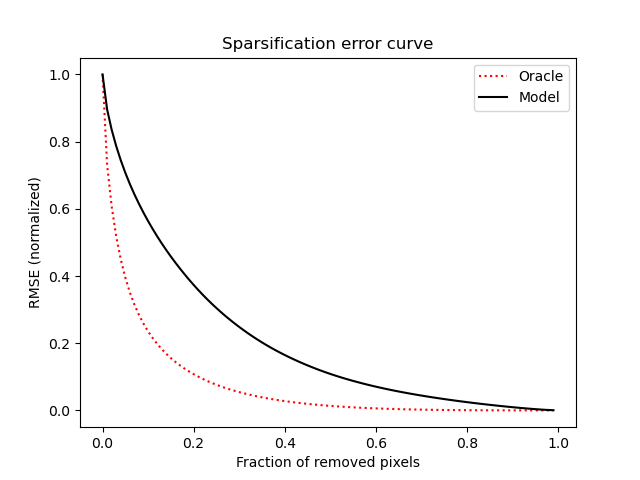}
  \caption{\scriptsize AUSE MCD CDec p=0.3 M=64}
\end{subfigure}
\caption{Calibration curves (AUCE and AUSE) for MC dropout and deep ensembles.}
\label{fig:u_metrics}
\end{figure}

Fig. \ref{fig:results} shows the evolution of the metrics with the number of forward passes $M$. 
We can take $M=1$ as the baseline, since there is no epistemic contribution. In general, predictions improve as $M$ increases. However, such improvements are hardly noticeable for $M>18$.
Interestingly, for MCD CDec, the epistemic uncertainty results in a worse uncertainty calibration in comparison to a single forward pass. 

Fig.~\ref{fig:u_metrics} shows the calibration error curves and sparsification error curves from where AUSE and AUCE were extracted. In these figures it can be seen that all models are overconfident, and the similarity between the sparsification error curves that leads to similar AUSE values.

\begin{figure}
    \centering
    \includegraphics[width=\columnwidth]{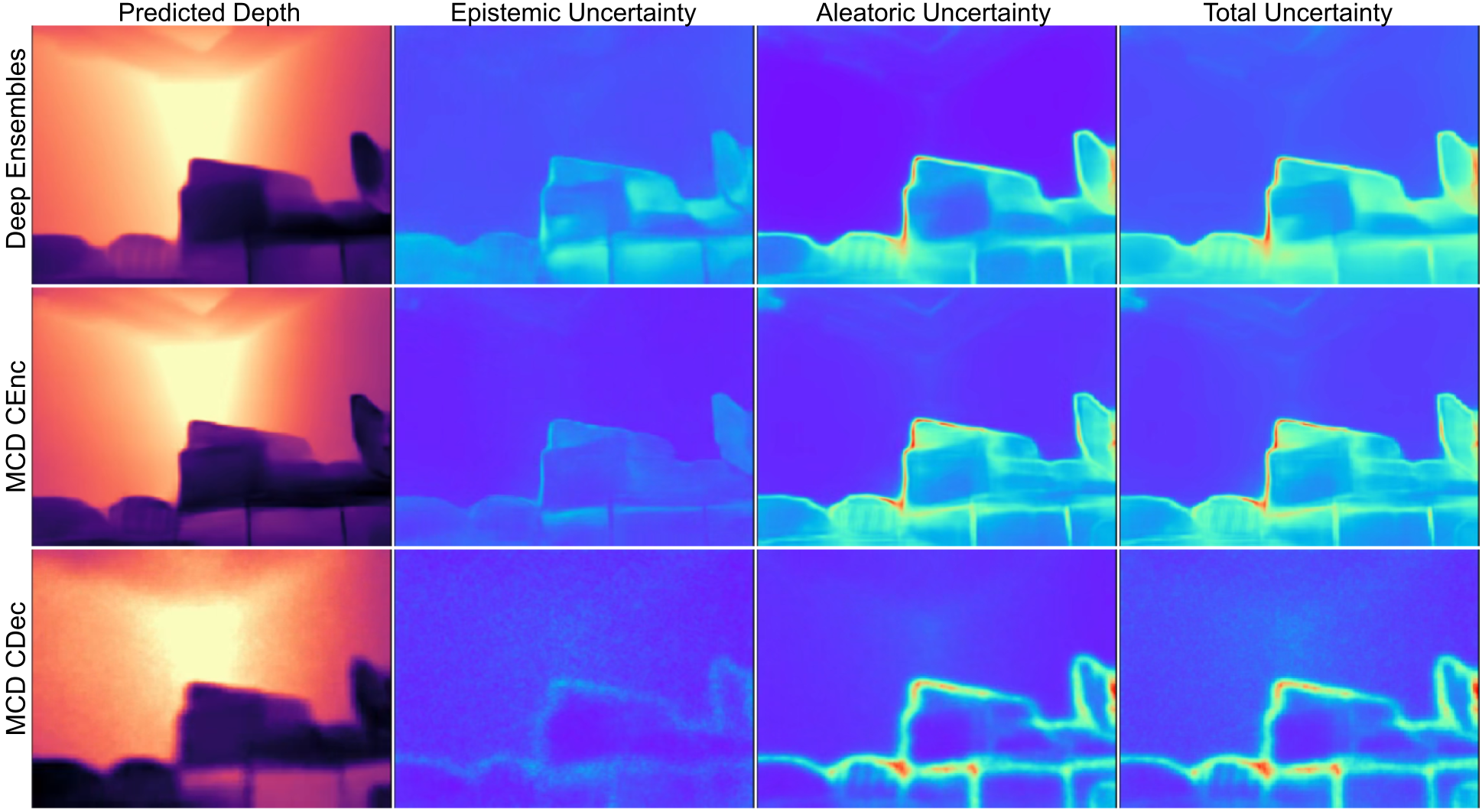}
    \caption{Depth and uncertainty predictions in a SceneNet RBG-D image.}
    \label{fig:qualitative_res}
\end{figure}

\begin{figure}
    \centering
    \includegraphics[width=\columnwidth]{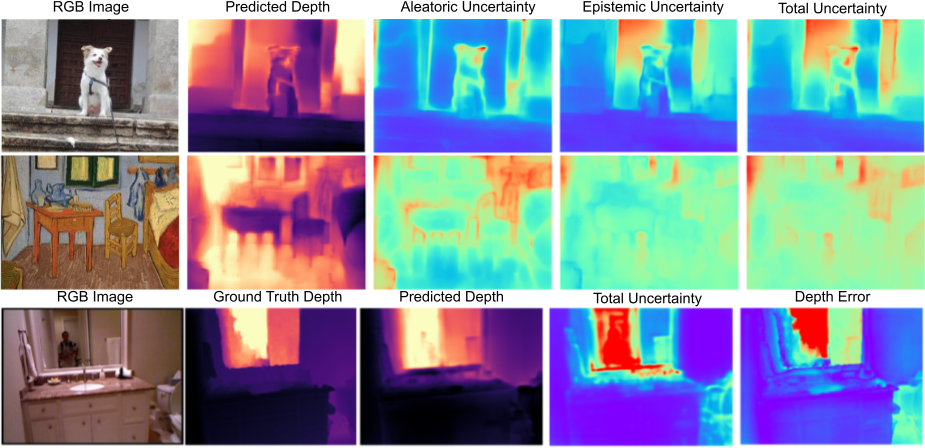}
    \caption{Predictions for three out-of-distribution images, showing high uncertainty for unfamiliar objects and textures. Top row: the aleatoric uncertainty is large in depth discontinuities and object boundaries, and the epistemic one concentrates in unknown patterns (the dog and the door). Middle row: aleatoric and epistemic uncertainties are both large due to large differences in appearance with respect to the training data. Bottom row: prediction in NYU Depth V2 image by an ensemble trained in SceneNet. The uncertainty is only very high for the human, not present in the training data.}
    \label{fig:outofdistr}
\end{figure}

\begin{table*}
\scriptsize

    \centering
    \begin{tabular}{|c|c||c|c|c|c|c|c|c||c|c|}
    \hline
        Model & M & Abs Rel  & Sq Rel & RMSE & RMSE Log & $\delta< 1.25$ & $\delta < 1.25^2$ & $\delta < 1.25^3$ & AUCE  & AUSE\\\hline
Deep Ensembles & 24 & \textbf{0.1431} & \textbf{0.1052} & \textbf{0.5842} & \textbf{0.1973} & \textbf{0.8157} & \textbf{0.9596 }&\textbf{ 0.9894} & \textbf{0.1302} &\textbf{ 0.1588}\\\hline

MCD CENC-CDEC p=0.3 & 32 & 0.1560 & 0.1250 & 0.6332 & 0.2155 & 0.7867 & 0.9454 & 0.9849 & 0.1453 & 0.1703\\\hline
MCD CDec p=0.3 & 32 & 0.1574 & 0.1286 & 0.6307 & 0.2160 & 0.7850 & 0.9460 & 0.9851 & 0.1382 & 0.1770\\\hline
MCD CEnc p=0.3 & 32 & \underline{0.1495} & \underline{0.1173} & \underline{0.6180} & \underline{0.2090} & \underline{0.8006} & \underline{0.9491} & \underline{0.9858} & 0.1436 & \underline{0.1653}\\\hline
MCD CENC-CDEC p=0.5 & 32 & 0.1642 & 0.1385 & 0.6592 & 0.2264 & 0.7671 & 0.9385 & 0.9824 & 0.1356 & 0.1668\\\hline
MCD CEnc p=0.5 & 32 & 0.1525 & 0.1220 & 0.6239 & 0.2127 & 0.7942 & 0.9467 & 0.9848 & 0.1377 & 0.1720\\\hline
MCD CDec p=0.5 & 32 & 0.1578 & 0.1291 & 0.6326 & 0.2164 & 0.7861 & 0.9457 & 0.9843 & \textbf{0.1286} & 0.1716\\\hline

    \end{tabular}
    \caption{Depth and uncertainty metrics for several variations of MC Dropout and Deep Ensembles in NYU Depth V2. Best results are boldfaced, second best ones are underlined.}
    \label{tab:nyumetrics}
\end{table*}

\begin{figure*}[ht!]
    \centering
    \includegraphics[width=\textwidth]{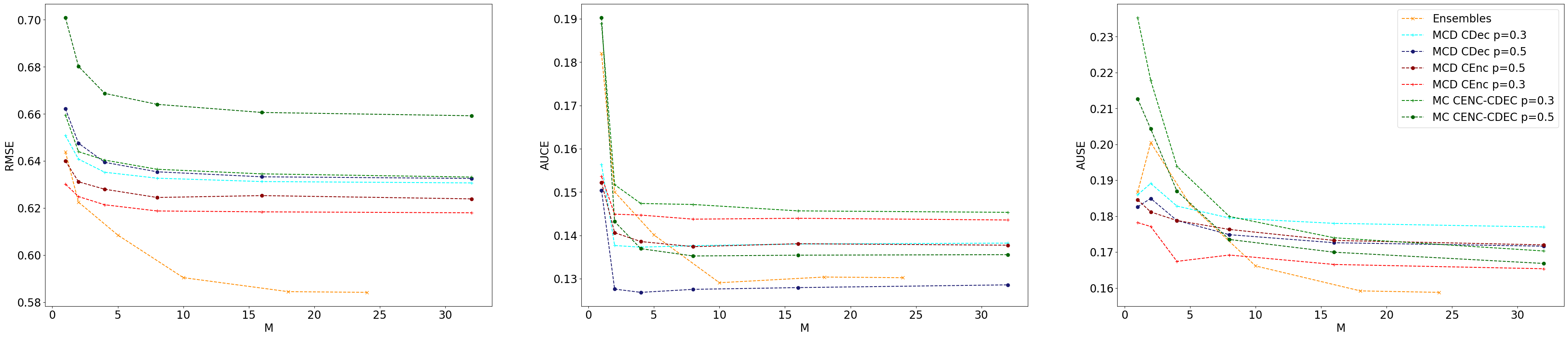}
    \caption{
Comparison of MCD CDec, CEnc and Deep Ensembles for forward passes $M$. Left: RMSE. Center: AUSE. Right: AUCE. The higher $M$ is, the better the performance, but with slight improvements for $M>18$.} 
    \label{fig:resultsnyu}
\end{figure*}

\begin{figure}
\scriptsize
    \centering
    \includegraphics[width=\columnwidth]{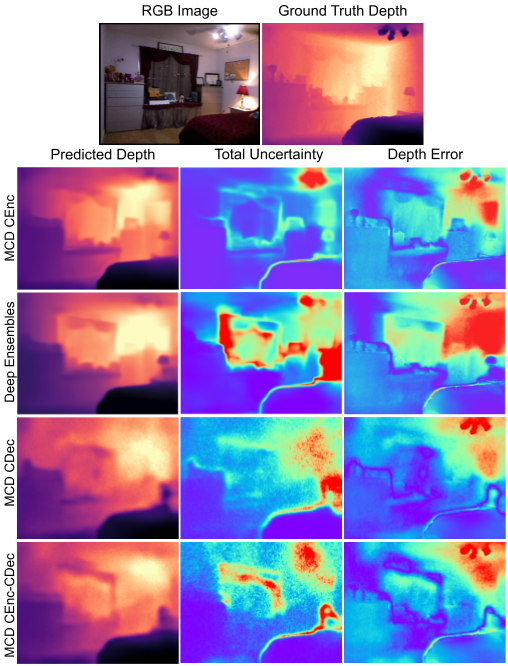}
    \caption{
 Depth and uncertainty results in NYU Depth v2 for MC dropout variations and deep ensembles. Top Row: input image and ground truth depth. First column: predicted depth. Second column: predicted uncertainty. Third column: depth error. Colors are equalized per image for better visualization.}
    \label{fig:qualitativeNYU}
\end{figure}

Table \ref{tab:nyumetrics} shows the metrics for depth and uncertainty on NYU Depth v2. We evaluate three MC dropout variations: MCD CEnc, MCD CDec, and MCD CEnc-CDec. In this dataset, deep ensembles show the best results in terms of uncertainty and depth metrics. 
As in SceneNet, applying dropout in the complete encoder (MCD CEnc) shows better results than applying it in the decoder (MCD CDec) or in the whole network (MCD CEnc-CDec). In this last case, the worse performance is caused by a too strong regularization effect. 
It is worth mentioning that the network is unable to recover the object boundaries and that the aleatoric uncertainty becomes diffuse for the high dropout rate case $p=0.5$. This effect becomes more noticeable when the decoder contains dropout layers. 
 Fig. \ref{fig:qualitativeNYU} shows results for one sample image. Observe that the areas with higher error correspond with the areas with higher uncertainty.

For all methods, increasing the number $M$ of forward passes improves the performance in uncertainty and depth (see Fig. \ref{fig:resultsnyu}). We do not see a noticeable improvement, though, for $M>18$ for deep ensembles and $M>32$ for MC Dropout.
Aleatoric uncertainty appears mainly in depth discontinuities, at object edges, and regions with sharp contrast in lighting (see Fig.~\ref{fig:qualitative_res}). As additional examples, Fig.~\ref{fig:outofdistr} shows highly uncertain depth predictions for three out-of-distribution images. It indicates high uncertainty values for unfamiliar objects not seen during training, like the dog and the door in the background of the first picture, the unrealistic patterns of the painting ``Bedroom in Arles'' by Van Gogh in the second one, and the person in the third one.

\subsection{Bayesian Pseudo-RGBD ICP}
\label{sec:pseudorgbdicp}

In this section, we evaluate the application of Bayesian depth neural networks for two-view relative motion. Relative motion can be directly computed from deep neural networks where the inductive bias can be used to estimate absolute values even for uncalibrated cameras and blurry or poorly illuminated images. But geometric methods are more precise in a multi-view setting with a known baseline for scale disambiguation and sensible features can be tracked between multiple views ~\cite{zhou2020learn}. In this work, we present a mixed approach where we use a two-view geometric method (ICP) to accurately compute the relative motion based on the depth prediction from neural networks. Thus, our approach is able to work with poor quality images and provide an unambiguous 3D transformation.


Our proposal leverages the depth predicted by a network to augment monocular images into what we call pseudo-RGBD views, and then aligns them using Iterative Closest Point (ICP). Similar ideas were proposed recently in \cite{tiwari2020pseudo,luo2020consistent}. Differently from us, they rely on Structure from Motion~\cite{schonberger2016structure} or visual SLAM~\cite{mur2017orb} to estimate the motion from the pseudo-RGBD views. Also, we use depth uncertainty for a more informed point cloud alignment, specifically excluding highly uncertain points from ICP.

\begin{figure}
    \centering
    \includegraphics[width=\linewidth]{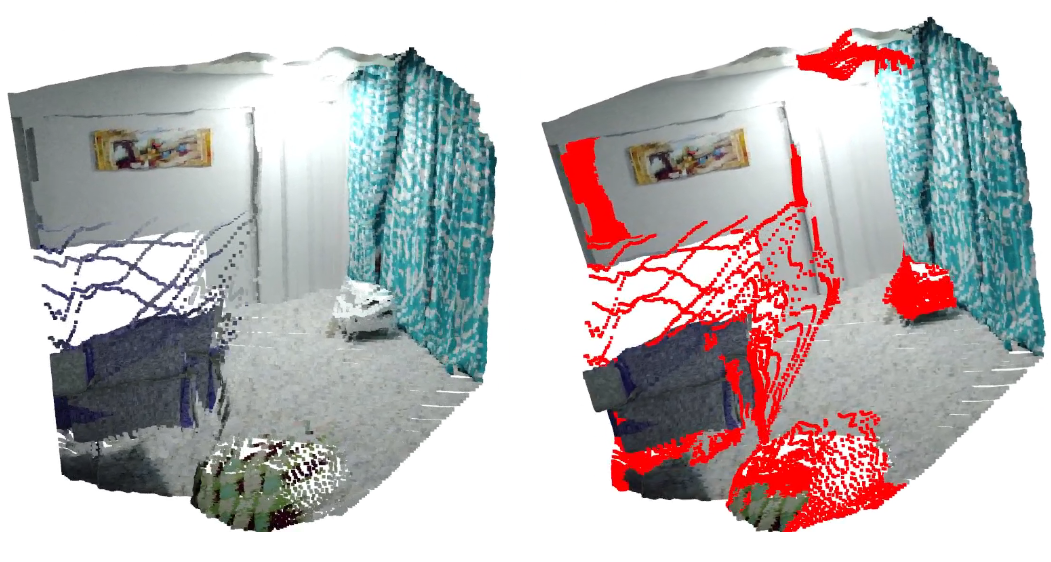}
    \caption{Left: Single-view reconstruction. Right: Same reconstruction, with the $10\%$ most uncertain points plotted in red. These points corresponds to spurious or high error points that will degrade the performance of ICP.}
    \label{fig:application}
\end{figure}


\begin{table}[h!]
\centering
\scriptsize
\begin{tabular}{ |p{1.27cm}|l|l|l|l|l|l|l| }
\hline
\textbf{Percentile} & .30 & .50 & .75 & .90 & .95 & .99 & 1.00\\\hline
\textbf{RMSE t} [m] & 0.238 & 0.216 & 0.188 & \underline{0.182} & \textbf{0.179} &0.190 &0.190\\\hline
\textbf{RMSE r} [$^\circ$] & 1.992 & 1.911 & 1.936 & \textbf{1.847} & \underline{1.888} &1.912 & 1.891\\

\hline
\end{tabular}
    \caption{ICP errors for percentiles .30, .50, .75, .90, .95, .99, 1.00. Best are boldfaced, second best underlined.}
    \label{tab:application}
\end{table}

Our experimental setup is as follows. We selected $1408$ random image pairs, separated by at least $4$ frames, from SceneNet RGB-D. 
We excluded pairs with large areas without ground-truth depth (e.g., windows) and small overlap (rotations larger than $60^\circ$). We kept the image pairs for which there is sufficient evidence that ICP converged.

For each pair, we back-projected the estimated depth distributions into a point clouds and applied ICP to the following percentiles of the most certain points according to our estimation: $.30$, $.50$, $.75$, $.90$, $.95$, $.99$, and $1.00$ (the percentile $1.00$ corresponds to the full point clouds). We used our deep ensemble model, as it showed the best uncertainty calibration in Table \ref{tab:metrics}.

Fig. \ref{fig:application} illustrates our hypothesis with an example. The left point cloud is the original one, and the right one corresponds to the percentile $.90$. The points highlighted in red represent the $10\%$ most uncertain points according to our uncertainty estimation, and clearly correspond to highly erroneous ones, as they lie on depth discontinuities. Removing such points from ICP will improve its accuracy.

Table \ref{tab:application} shows the translational and rotational errors for all evaluated pairs. As motivated before, removing the most uncertain points (corresponding in well calibrated models to the most errouneous ones) reduces the estimation errors. The best results are for percentiles $.90$ and $.95$ ($1.00$ corresponds to the original point cloud).
When the percentage of points removed is higher the error grows. This effect becomes obvious for percentiles $.30$ and $.50$, for which $70\%$ and $50\%$ of the points with the highest uncertainty were removed respectively. In these cases, the number of points removed is too large and the estimation becomes less accurate.

\section{Conclusions}
In this paper, we evaluated MC dropout and deep ensembles as scalable Bayesian approaches to uncertainty quantification for single-view supervised depth learning.
We demonstrate empirically that using MC dropout in the encoder outperforms other variations used in the literature, which is a result of practical relevance. The placement of dropout in the architecture indeed has a significant effect in the estimation of depth and uncertainty. We believe that this finding would be also relevant for works based on noise layers like \cite{postels}. As a second conclusion of our analysis, deep ensembles have the best calibrated uncertainty estimations. However, applying dropout in the encoder performs only slightly worse than deep ensembles. 
As MC dropout needs much less memory than deep ensembles, it may be considered the Bayesian approach with more potential for applications.

In our experimental results, we also show the application of Bayesian depth networks to pseudo-RGBD ICP, with the result that relative transformation can be improved by excluding the points with higher uncertainties. For future investigation into this topic, we would like to address the overconfidence that we observed in our depth prediction networks and research a better representation of uncertainty. Furthermore, it would be interesting to investigate how to benefit from the posterior computation to overcome the domain changes from synthetic to real data and between different scenes. 

\balance

{\small
\bibliographystyle{ieee}
\bibliography{egbib}
}

\end{document}